%% file: CVPR2027-main-2165.tex
\documentclass[10pt,twocolumn,letterpaper]{article}

\usepackage[pagenumbers]{cvpr} 

\input{preamble}

\usepackage{marvosym}

\definecolor{cvprblue}{rgb}{0.21,0.49,0.74}
\usepackage[pagebackref,breaklinks,colorlinks,allcolors=cvprblue]{hyperref}

\definecolor{Gray}{gray}{0.9}
\definecolor{MetricBlue}{RGB}{146,176,223}
\newcommand{\hzw}[1]{\textcolor{black}{#1}}

\def\confName{CVPR}
\def\confYear{2027}

\title{Beyond a single frame: multi-frame reflection removal}

\author{
	Yu Chen\textsuperscript{1,2}\quad
	Xi Tong\textsuperscript{1,2}\quad
	Jinlei Li\textsuperscript{1,2}\quad \\
	Chunyu Zhu\textsuperscript{5}\quad 
	Zixuan Chen\textsuperscript{4}\quad
	Du Chen\textsuperscript{3}\quad
	Zhe-Ming Lu\textsuperscript{1,2}\quad 
	Zewei He\textsuperscript{1,2}\textsuperscript{\Letter} \\
	\textsuperscript{1}Zhejiang University\quad
	\textsuperscript{2}Huanjiang Laboratory \quad 
	\textsuperscript{3}Vivo \quad \\
	\textsuperscript{4}The Chinese University of Hong Kong \quad 
	\textsuperscript{5}Xi'an University of Electronic Science and Technology \\
}

\begin{document}
\twocolumn[{
	\maketitle
	\begin{center}
		\captionsetup{type=figure}
		\includegraphics[width=0.95\textwidth]{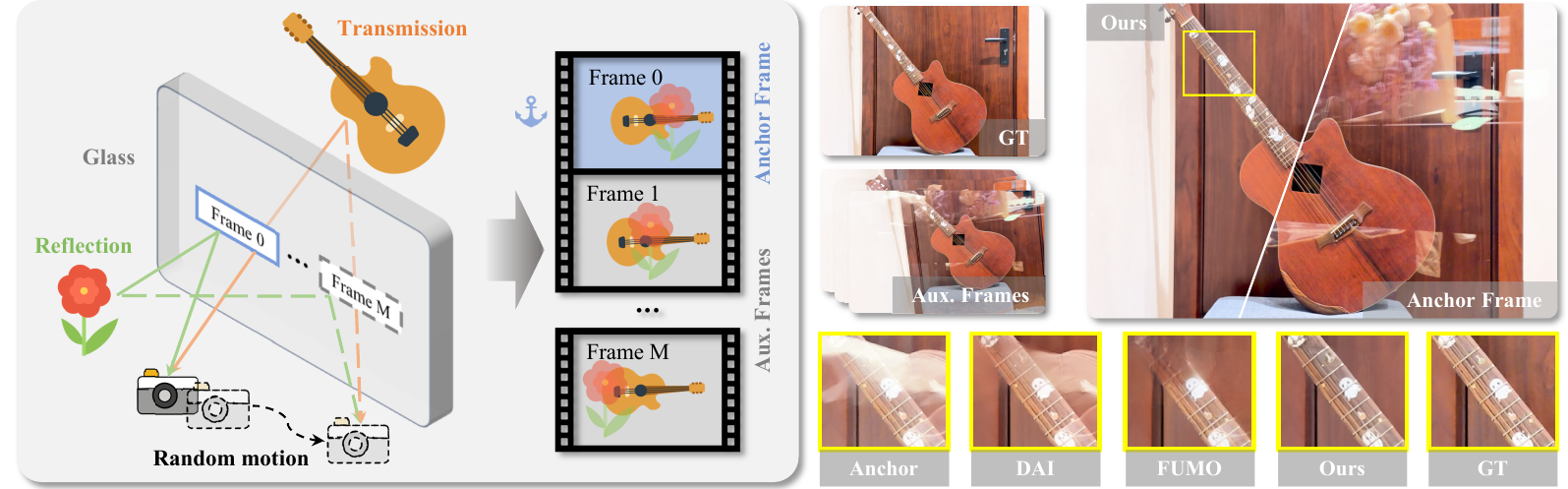}
		\vspace{-5pt}
		\captionof{figure}{\textbf{Left}: Problem setup of multi-frame reflection removal (MFRR). A sequence of auxiliary images are captured to help remove the reflection of the anchor frame. It represents a typical anchor-centric task in computational imaging. \textbf{Right}: Comparative results with several single-image reflection removal methods.}
		\label{fig1-teaser}
	\end{center}
}]

\begin{abstract}
Multi-frame reflection removal (MFRR) offers a promising research direction by exploiting complementary information across different viewpoints.
However, it remains underexplored due to the scarcity of training data, complex cross-frame dependencies, and interference from local motion.
In this paper, we attempt to address these issues.
First of all, we introduce MulVi, the pioneering dataset specifically designed for MFRR, consisting of a training set generated through 3D scene-based rendering and real-world benchmarks for comprehensive evaluation.
Building upon MulVi, we propose MFRR-DiT, an end-to-end Diffusion Transformer framework that formulates MFRR as a multi-image-conditioned generation task, enabling cross-view information aggregation without explicit geometric alignment.
To better exploit the anchor-centric nature of MFRR, we develop Anchor-Centric Attention (ACA), which regulates cross-frame interactions to facilitate complementary information aggregation while mitigating interference from local motion. 
Furthermore, we introduce a test-time frame sampler that accommodates a variable number of candidate frames and selects informative and complementary auxiliary observations. 
Extensive experiments demonstrate that MFRR-DiT outperforms existing reflection removal methods on challenging real-world benchmarks.
\end{abstract}
\input{sec/1_intro}

\input{sec/4_method}

\input{sec/5_experiment}

\input{sec/6_conclusion}

\newpage
\section*{LLM Usage Statement}
We used ChatGPT solely for English language polishing and grammatical refinement during manuscript preparation. 
No LLMs were used for research ideation, methodology development, experimental design, data analysis, or generation of scientific content.
All LLM-assisted text was reviewed and revised by the authors, who take full responsibility for the accuracy, originality, and integrity of the final manuscript.
{
    \small
    \bibliographystyle{ieeenat_fullname}
    \bibliography{main}
}

\end{document}

%% file: preamble.tex
\usepackage{algorithm}
\usepackage{algorithmic}

\usepackage{amsfonts}
\usepackage{color}
\usepackage{adjustbox}
\usepackage{booktabs}
\usepackage{multirow}
\usepackage{colortbl}
\usepackage{amssymb}
\usepackage{amsmath}

\usepackage{graphicx}
\usepackage{caption}
\usepackage{lipsum}

\usepackage{fancyhdr}

%% file: sec/1_intro.tex
\section{Introduction}
\label{sec: intro}
Reflection removal aims to recover the underlying transmission layer from images captured through reflective mediums, such as glass or transparent plastic \cite{Hu2021NeurIPS-YTMT,Hu2023ICCV-DSRNet,Hu2024NeurIPS-DSIT,Zhao2025CVPR-RDNet,Huang2025TPAMI-DExNet,Hu2026AAAI-DAI,Chen2026CVPR-GFRRN,Huang2026CVPR,Li2026CVPR-GenSIRR,Lee2026CVPR-ReflexSplit}. 
The observed image represents a complicated mixture of transmitted and reflected components, making the restoration inherently ambiguous without additional observations or prior knowledge.
With the widespread use of smartphone cameras, reflection has become a common issue in daily photography, from capturing exhibits in museum display cases and taking photos through car windows to photographing outdoor scenes through windows \cite{Niklaus2021WACV,Xu2026ECCV,Liu2026ECCV-DiffUR3,He2026ECCVW}.
Reflections in these scenarios significantly degrade image quality, highlighting the urgent need for effective reflection removal methods.

Single-image reflection removal (SIRR) is the most widely studied setup.
Despite recent progress in SIRR, existing state-of-the-art methods \cite{Zhao2025CVPR-RDNet,Hu2026AAAI-DAI,Chen2026CVPR-GFRRN,Huang2026CVPR,Li2026CVPR-GenSIRR,Lee2026CVPR-ReflexSplit} still struggle with challenging cases where transmission and reflection are difficult even for human beings to distinguish at a glance (see in the right part of Fig.~\ref{fig1-teaser}). 

This ambiguity has motivated researchers to explore various sources of auxiliary information for separating transmission and reflection, including polarization cues \cite{Yao2025CVPR-PolarFree,Lyu2019NeurIPS,Lyu2023TPAMI}, event streams \cite{Wang2026ICML-event}, near-infrared images \cite{Hong2020ICME-infrared,Hong2023TMM-infrared}, and flash/no-flash pairs \cite{Chang2020IJCV-SDN,Lei2021CVPR-flash-only,Lei2023TPAMI-flash-only,Hong2024IJCV}.
However, these approaches often impose additional hardware requirements or capture constraints, limiting their practical applicability. 
Polarization-, event-, and near-infrared-based methods typically require specialized imaging devices, some of which can be costly.
Active near-infrared imaging used in \cite{Hong2020ICME-infrared,Hong2023TMM-infrared} is constrained by its effective operating range, restricting the practicability in outdoor scenarios.
Flash-based methods require careful control over the illumination direction, and they are unsuitable for locations where flash photography is prohibited, such as many museums.

Another line of research adopts a multi-view setting, which takes as input an image sequence captured by deliberately moving the camera \cite{Li2021TIP,Guo2014CVPR,Li2013ICCV}.
We denote this direction as multi-frame reflection removal (MFRR).
The fundamental assumption of MFRR is that complementary information from other viewpoints can aid reflection removal. 
Multi-view methods generally require the set of images to be captured with a large baseline between them.
They typically build upon complex inter-frame alignment or optimization, significantly increasing both computational burden and user effort \cite{Guo2014CVPR,Li2013ICCV}.
A simplified and user-friendly multi-view setting is dual-view, leveraging complementary observations from only two viewpoints to facilitate the separation of transmission and reflection layers \cite{Niklaus2021WACV,Yu2025ACMMM-GausSplat,Xu2026ECCV}.
Compared with multi-view, the dual-view setting has access to fewer complementary information, which may limit its theoretical upper bound on performance.

Despite the considerable potential of MFRR, this direction remains relatively underexplored in the deep learning era due to three major challenges:
(1) \textbf{Lack of data}: 
For MFRR, each training sample should comprise a reflection-contaminated anchor frame, a sequence of reflection-contaminated frames from different viewpoints, and a reflection-free ground truth image pixel-wise aligned with the anchor frame.
Learning-based methods require a substantial number of such pairs. 
However, existing multi-view datasets are very limited in size \cite{Chugunov2024CVPR-NSF,Xue2015TOG,Li2013ICCV} (primarily designed for evaluation only) or not fully open-source yet \cite{Niklaus2021WACV,Xu2026ECCV}. 
(2) \textbf{Complex cross-frame dependencies}:
The complexity arises from two aspects. 
First, the number of input frames is not fixed and can be variable, which introduces a fundamental trade-off between information sufficiency and redundancy. While a small number of frames may provide insufficient complementary cues, a large number of frames can introduce substantial redundancy and computational overhead.
Furthermore, viewpoint changes cause spatial misalignment across frames, making reliable correspondence estimation and feature aggregation particularly challenging, especially under severe reflection interference.
(3) \textbf{Local motion interference}: 
Beyond the global motion induced by camera movement, objects in both the transmission and reflection layers may undergo independent local motions. 
Such entangled motion patterns introduce an additional challenge unique to MFRR, making it difficult to distinguish genuine scene content (\ie, transmissions) from unwanted reflections.

In this paper, we revisit MFRR from an anchor-centric perspective and propose an end-to-end framework based on Diffusion Transformer (MFRR-DiT). 
Instead of explicitly estimating cross-frame correspondences or imposing geometric constraints, \textit{we formulate MFRR as a multi-image-conditioned generation task}, allowing complementary observations from different viewpoints to be integrated directly within the transformer.
To the best of our knowledge, this is the first diffusion-based generation framework for general MFRR.

In this framework, we develop Anchor-Centric Attention (ACA), a task-driven attention mechanism that exploits the asymmetric restoration objective of MFRR. 
\hzw{By jointly regulating cross-frame information flow and positional representations, ACA facilitates effective information aggregation from complementary viewpoints while suppressing unnecessary interactions.}
Empirically, we find that this anchor-centric attention scheme helps mitigate interference caused by local motion in auxiliary frames.
Furthermore, we develop a test-time frame sampler with a two-step selection strategy that accommodates a variable number of candidate frames and selects auxiliary observations that are informative with respect to the anchor and complementary to one another.

At last, to support the development and evaluation of learning-based MFRR methods, we additionally construct MulVi, a dataset comprising large-scale rendered training sequences and real-world evaluation benchmarks.

\section{Related work}
\label{sec: related work}
\subsection{Single-image reflection removal}
Learning-based methods have made substantial progress in
single-image reflection removal. Zhang \etal~\cite{Zhang2018CVPR} train a fully convolutional
network with perceptual and exclusion losses.
Dual-stream methods~\cite{Hu2021NeurIPS-YTMT,Hu2023ICCV-DSRNet,Hu2024NeurIPS-DSIT,Chen2026CVPR-GFRRN,Lee2026CVPR-ReflexSplit}
model transmission and reflection through feature interaction.
Recent approaches~\cite{Hu2026AAAI-DAI,Li2026CVPR-GenSIRR,Xu2026ECCV-FUMO,Shin2026ECCV-PRISM,Huang2026CVPR}
exploit pretrained generative priors for reflection removal
or joint layer separation.
However, single-image separation remains ambiguous,
especially under strong reflections.

\subsection{Multi-frame reflection removal}
Multiple observations can provide complementary appearance
and geometric cues.
Early methods exploit cross-frame transmission correlation
and gradient sparsity and independence
priors~\cite{Guo2014CVPR}, or jointly estimate layer appearances
and motions from parallax~\cite{Xue2015TOG}.
For dual-view inputs, Niklaus \etal~\cite{Niklaus2021WACV}
align transmission content using reflection-invariant optical
flow, while Xu \etal~\cite{Xu2026ECCV} combine cross-view
correlations and camera-pose conditioning in a diffusion
framework.
Liu \etal~\cite{Liu2020CVPR-learning} alternate layer-specific
flow estimation and neural reconstruction.
Li \etal~\cite{Li2021TIP} use multi-view edge-depth estimates
to classify layer edges and guide transmission reconstruction,
while Prasad \etal~\cite{Prasad2023WACV-Burst} derive reflection
motion aggregation cues from handheld bursts.
Related layer-separation approaches include
Visual Centrifuge~\cite{Alayrac2019CVPR-visual_centrifuge},
which decomposes blended videos using 3D convolutions,
and Neural Spline Fields~\cite{Chugunov2024CVPR-NSF},
which jointly optimize image and motion representations
at test time for burst fusion and layer separation.

\subsection{Multimodal reflection removal}
Multimodal approaches provide complementary cues for
reflection removal.
PolarFree~\cite{Yao2025CVPR-PolarFree} combines polarization
with diffusion-generated priors.
Hong \etal~\cite{Hong2020ICME-infrared} exploit weaker
reflections in active near-infrared observations under
suitable capture conditions.
EvReflection~\cite{Wang2026ICML-event} uses event-derived
micro-dynamics as layer-specific motion cues.
Lei and Chen~\cite{Lei2021CVPR-flash-only} derive flash-only
cues by subtracting paired ambient images from flash images
in the raw domain.

\begin{figure}[t]
	\centering
	\includegraphics[width=0.99\linewidth]{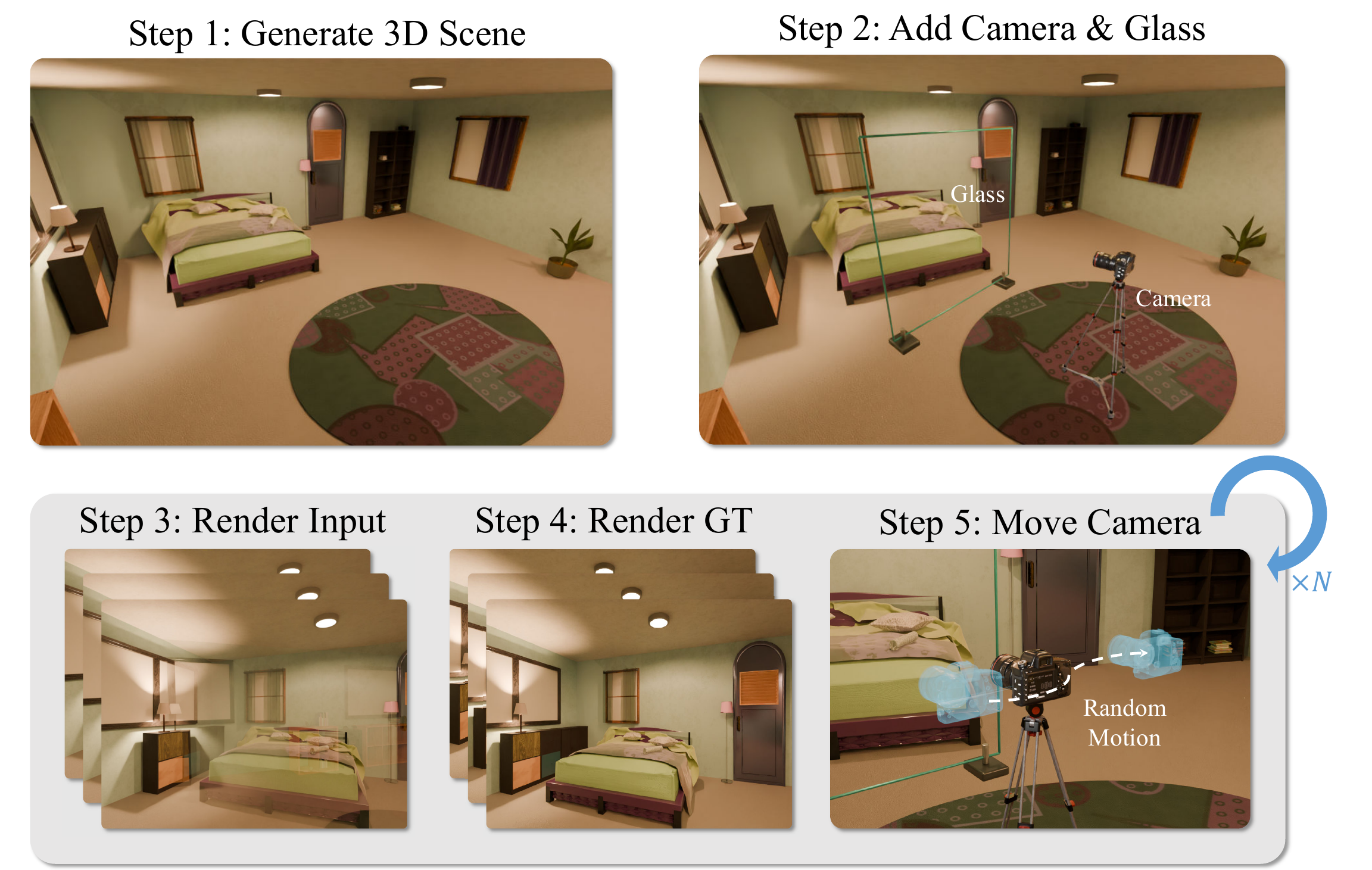}
	\vspace{-5pt}
	\caption{The data generation pipeline of our MulVi-training dataset. We adopt the synthetic 3D scenes to render the image sequences.}
	\vspace{-10pt}
	\label{fig:fig2}
\end{figure}

\section{MulVi dataset}
\label{sec: dataset}

\begin{figure}[t]
	\centering
	\includegraphics[width=0.99\linewidth]{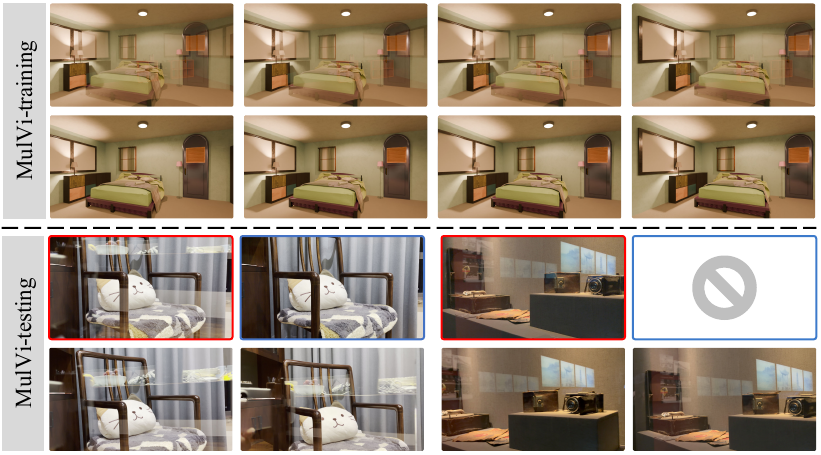}
	\vspace{-5pt}
	\caption{Visual examples from our MulVi dataset. \textbf{Top}: rendered multi-view sequences from MulVi-training, with corresponding reflection-free GTs. \textbf{Bottom}: Real-captured sequences from MulVi-testing, including controlled scenes with GTs and in-the-wild scenes without GTs. {\color{red}Red} and {\color{blue}blue} borders indicate the anchor frames and their corresponding reflection-free GTs, respectively.}
	\vspace{-10pt}
	\label{fig:fig3}
\end{figure}

Existing reflection removal datasets are mainly collected for SIRR \cite{Zhu2024CVPR-RRW,Chen2026CVPR-GFRRN}.
Then, conventional multi-view datasets are originally designed solely for evaluation and are limited in scale \cite{Chugunov2024CVPR-NSF,Xue2015TOG,Li2013ICCV}.
The aforementioned datasets are unsuitable for training MFRR models, as each training sample requires a reflection-contaminated anchor frame, additional reflection-contaminated frames from different viewpoints, and a reflection-free ground truth image pixel-wise aligned with the anchor frame.
In addition, possibly due to commercial licensing and copyright restrictions, current dual-view datasets are not publicly available \cite{Niklaus2021WACV,Xu2026ECCV}.
\hzw{To bridge this gap, we propose a \textbf{Mul}ti-\textbf{Vi}ew dataset (\ie, \textbf{MulVi}), the first MFRR	dataset that satisfies all the aforementioned requirements.}
Our MulVi dataset contains two parts: (1) MulVi-training: 2,255 rendered sequences with corresponding reflection-free ground truths; and (2) MulVi-testing: 50 real-captured pairs in controlled environments with ground truths, and 100 real-captured pairs in the wild without ground truths.
We elaborate on the details of the MulVi dataset as follows.

\begin{figure*}[t]
	\centering
	\includegraphics[width=0.99\linewidth]{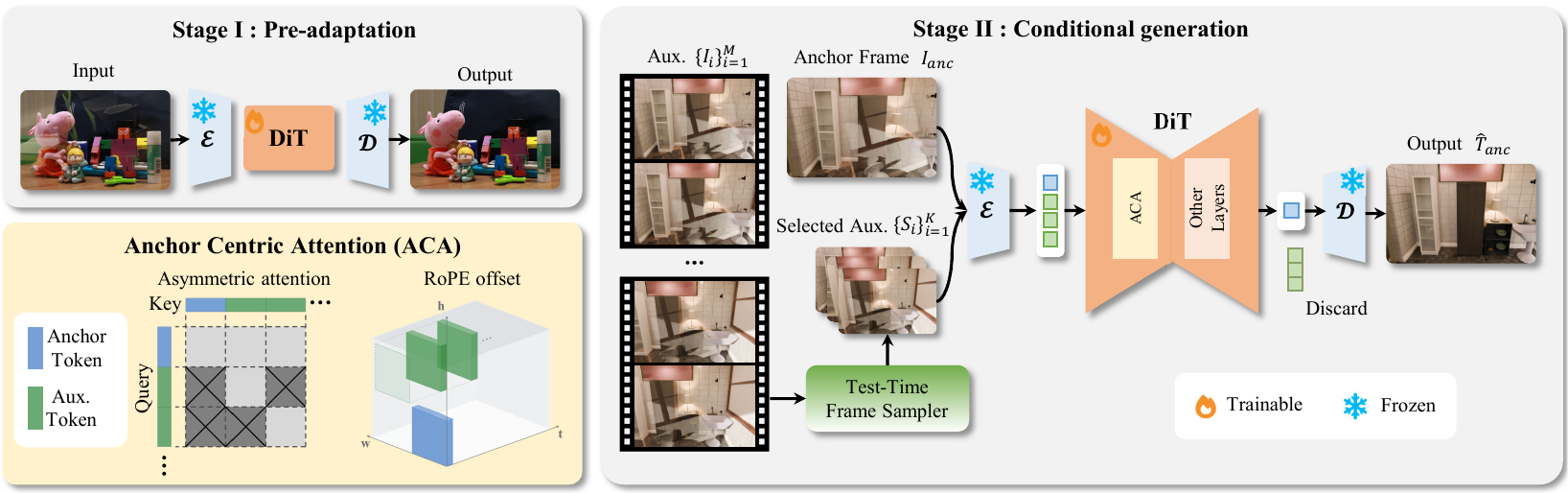}
	\vspace{-5pt}
	\caption{The details of our proposed end-to-end MFRR framework based on Diffusion Transformer, \ie, MFRR-DiT. It contains two training stages: (1) pre-adaptation stage and (2) conditional generation stage. An anchor-centric attention (ACA) is tailored to fit MFRR task. During inference, only Stage II is employed, with a test-time frame sampler selecting informative auxiliary frames.}
	\vspace{-10pt}
	\label{fig:fig4}
\end{figure*}

\subsection{Generation pipeline of MulVi-training}
Acquiring a large-scale real-world training dataset for MFRR task requires substantial time and resources.
Therefore, we propose a 3D scene-based rendering pipeline, as shown in Fig.~\ref{fig:fig2}.
First, we employ Infinigen \cite{Raistrick2023CVPR,Raistrick2024CVPR}, an open-source procedural scene generation framework, to construct 300 high-quality 3D scenes. 
Infinigen enables fully procedural generation of photo-realistic 3D scenes, where all assets, ranging from large-scale structures to fine-grained details, are generated algorithmically without relying on external static assets.
We then incorporate a virtual glass surface and a virtual camera into the generated 3D scenes to simulate realistic reflection-contaminated observations (see in the step 2 of Fig.~\ref{fig:fig2}).
\hzw{The virtual camera is positioned to observe a scene through the glass.
}
Next, we render the reflection-contaminated observation $I$ with Cycles, which is Blender’s ray-trace based production render engine.
The reflection-free ground truth image $T$ is rendered by setting specific hyper-parameters according to \cite{Zakarin2026CVPRW}.
Finally, we randomly adjust the camera pose and render the scene at each new viewpoint, repeating this process $N$ times to generate an image sequence, \ie, $\{I_i,T_i\}_{i=1}^{N}$.
Some visual examples are shown in Fig.~\ref{fig:fig3}.
In our implementation, we set $N=9$, and our MulVi-training dataset comprises a total of 40,590 frames.
During the training phase, we randomly sample $K+1$ frames from the generated sequence ($1\le K\le N-1$) and randomly designate one of them as the anchor frame, while treating the remaining $K$ frames as auxiliary frames.
The corresponding reflection-free image of the anchor frame serves as the ground truth.

Our 3D scene-based rendering pipeline fundamentally differs from the 2D image-based approach adopted in \cite{Zakarin2026CVPRW}.
The latter \cite{Zakarin2026CVPRW} assumes that objects in either the transmission or reflection layer lie on a single plane, which oversimplifies the underlying 3D geometry and fails to capture the geometric complexity of real-world scenes.

\hzw{Models trained solely on synthetic data often exhibit degraded performance on real-world images due to the domain gap between synthetic and real data 
\cite{Chen2024ECCV-PTTD,Chen2024TIP-DEA-Net,Mayer2018IJCV,Niklaus2021WACV}.
To cope with this issue, we also employ real-captured RRW dataset \cite{Zhu2024CVPR-RRW} to augment the training data.
Details can be found in Section~\ref{subsec: implementation}.
}

\subsection{MulVi-testing: real-world evaluation data}
To benchmark developed MFRR methods in terms of performance and generalization, we capture 50 pairs in controlled environments with corresponding labels, and 100 pairs in the wild without labels.

For the former, we first mount a camera or smartphone on a tripod and use a light-blocking box to eliminate reflections from the camera side, capturing a reflection-free ground truth image $T$. 
We then remove the box and capture a reflection-contaminated anchor frame $I_{anc}$ from the same viewpoint, ensuring spatial alignment with the ground truth. 
Finally, we detach the camera or smartphone from the tripod and record a video while moving it to obtain a sequence from $M$ different viewpoints.
Each testing pair can be expressed as $\{I_{anc},\{I_i\}_{i=1}^{M},T\}$.
For the latter, we first capture an anchor frame $I_{anc}$ and then record a frame sequence using a handheld camera or smartphone.
Each testing pair can be expressed as $\{I_{anc},\{I_i\}_{i=1}^{M},\varnothing\}$.
For our case, we set $M=39$.
Some visual examples are shown in Fig.~\ref{fig:fig3}.
We will publicly release the entire dataset upon acceptance to facilitate further research on MFRR.

MulVi-testing also covers challenging scenarios involving local motion, a common issue in practical multi-frame acquisition.
These scenarios provide a more realistic evaluation of MFRR methods under practical conditions.

%% file: sec/4_method.tex
\section{Methodology}
\label{sec: methodology}

\subsection{Problem definition}
Given a reflection-contaminated anchor frame $I_{anc}$ and $M$ auxiliary frames $\{I_i\}_{i=1}^{M}$ from different viewpoints, the goal of MFRR is to remove the reflection from $I_{anc}$ to estimate the underlying transmission layer $\hat{T}_{anc}$.
The operation of MFRR can be formulated as $\mathcal{F}_{MFRR}:\{I_{anc},\{I_i\}_{i=1}^{M}\} \rightarrow \hat{T}_{anc}$.
Here, $M$ denotes the number of auxiliary frames, with $M=1$ corresponding to the dual-view reflection removal setting.

Video reflection removal (VRR) \cite{Wang2026-S2R} is a very similar task to MFRR.
\hzw{Although both MFRR and VRR exploit information across multiple frames, they differ in their primary objectives. 
VRR aims to recover an entire reflection-free video sequence, with particular emphasis on temporal consistency across the restored frames. 
In contrast, MFRR focuses on reconstructing a single high-quality, reflection-free image at a designated anchor viewpoint.
}

\subsection{Our MFRR-DiT}
As shown in Fig.~\ref{fig:fig4}, we propose an end-to-end MFRR framework based on Diffusion Transformer (\ie, MFRR-DiT), which contains two training stages (Stage I: Pre-adaptation, and Stage II: Conditional generation). After training, only Stage II is employed for inference.

\begin{figure}[t]
	\centering
	\includegraphics[width=0.99\linewidth]{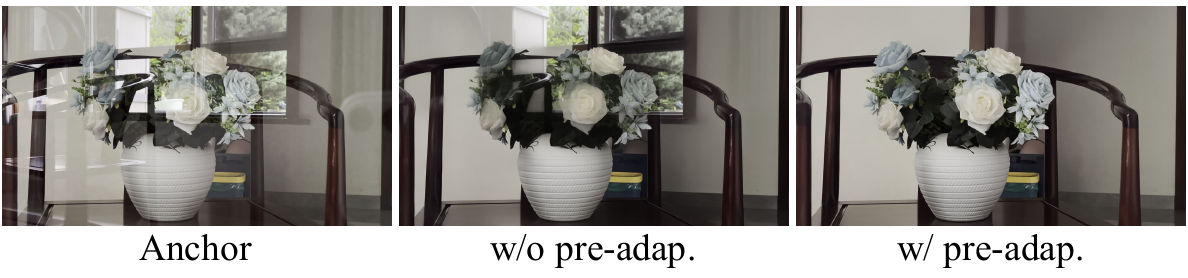}
	\vspace{-5pt}
	\caption{Experimental results with and without stage I. w/o pre-adap. means only stage II is trained.}
	\vspace{-10pt}
	\label{fig:fig5}
\end{figure}

\subsubsection{Stage I: Pre-adaptation}
Although pre-trained DiT models \cite{Peebles2023ICCV,Labs2025Arxiv-Flux} have powerful image generation capabilities, they are not explicitly optimized for reflection removal.
Directly introducing multi-frame conditions may therefore require the model to simultaneously learn reflection removal and complex cross-frame dependencies, making optimization more challenging.

To address this issue, we introduce a pre-adaptation stage to fine-tune the pre-trained DiT model (\ie, FLUX.2 Klein 4B) with paired reflection-contaminated and reflection-free images.
The training is flow matching-based by following \cite{Shin2026ECCV-PRISM} without estimating the reflection.
After that, the fine-tuned DiT obtains the reflection removal capabilities before extending it to multi-frame conditional generation.

As shown in Fig.~\ref{fig:fig5}, omitting the pre-adaptation stage leads to a substantial performance degradation, demonstrating the effectiveness of two-stage training.

\subsubsection{Stage II: Conditional generation}
\noindent\textbf{Multi-frame conditioning.}
Then, we extend the fine-tuned DiT to multi-image-conditioned generation.
We first encode the anchor frame and the auxiliary frames into the latent space using the pre-trained VAE encoder. 
The resulting latents are converted into image tokens and jointly fed into the DiT, allowing information from different viewpoints to be integrated during the generation process.
Since our objective is to recover the transmission layer at the anchor viewpoint, only the generated target tokens are retained and decoded to obtain the reflection-free image $\hat{T}_{anc}$.

\noindent\textbf{Anchor-centric attention.}
A straightforward strategy for multi-frame conditioning is to concatenate all image tokens and apply conventional self-attention. 
However, this introduces unrestricted interactions among auxiliary frames, which may be unnecessary for our anchor-centric generation objective.

To facilitate effective cross-frame information aggregation while suppressing unnecessary interactions, we propose an Anchor-Centric Attention (ACA) mechanism (see in Fig. \ref{fig:fig4}). 
\hzw{Unlike conventional self-attention, ACA establishes an asymmetric attention pattern tailored to the anchor-centric nature of MFRR.
Specifically, anchor tokens are allowed to attend to all auxiliary frames, whereas each auxiliary frame is restricted to attending to itself, preventing information flow from the anchor to auxiliary frames and direct information exchange among different auxiliary frames.}
\hzw{This design offers an additional advantage by reducing the theoretical computational complexity of self-attention.}

In addition, FLUX.2 adopts the RoPE mechanism \cite{Su2024-RoPE} to encode spatial information for both anchor and auxiliary tokens.
Originally, their position indices are identical, and it works well for spatially aligned situations \cite{Tan2025ICCV-OminiControl}.
Following \cite{Tan2025ICCV-OminiControl,Xue2026CVPR-Stand-In,Zhang2025ICCV-EasyControl}, we shift the position indices of the auxiliary tokens by a fixed offset, ensuring no spatial overlap.

\hzw{The proposed ACA also helps mitigate interference caused by local motion in auxiliary frames. 
Specifically, restricting direct interactions among auxiliary frames reduces the potential propagation of motion-inconsistent information.
More discussions can be found in Section~\ref{subsec: ablation}.
}

\subsection{Test-time frame sampler}
In real-world scenarios, the number of available auxiliary frames $M$ is not fixed, requiring our MFRR-DiT model to accommodate a variable number of inputs.
Therefore, we introduce a frame sampler with specific selection strategy to identify $K$ informative frames.
To select the most informative auxiliary frames, the selection strategy should satisfy the following three criteria:
\textbf{First}, the selected auxiliary frames should maintain enough scene overlap to find sufficient feature correspondences with the anchor frame, ensuring that they observe the same underlying transmission scene.
\textbf{Second}, the selected auxiliary frames should exhibit sufficient baseline from the anchor frame to ensure adequate cross-view variations.
\textbf{Third}, redundancy among the selected auxiliary frames should be minimized to ensure that each frame contributes useful complementary information.

Accordingly, our frame sampler adopts a two-step selection strategy that jointly considers the informativeness of individual auxiliary frames (to the anchor) and their inter-frame complementarity.
For clarity, we explain the selection process with identifying two auxiliary frames, while its extension to multiple auxiliary frames is straightforward.
\textbf{The pseudo code of our auxiliary frame selection strategy can be found in our supplementary material.}

\subsubsection{Anchor-aware selection}
In the first step, we evaluate the potential contribution of each candidate auxiliary frame $\{I_i\}_{i=1}^{M}$ to the anchor frame $I_{anc}$. 
To reduce computational overhead, all input images are first resized (longer side to 256) while preserving their aspect ratios.
For each auxiliary frame $I_i$, we estimate a homography $H_{i\rightarrow anc}$ that aligns it with the anchor frame $I_{anc}$. 
We then measure their geometric overlap as:
\begin{equation}
	o_i=\frac{|\Omega_{anc}\cap H_{i\rightarrow anc}(\Omega_i)|}{|\Omega_{anc}|},
	\label{Eq:Eq1}
\end{equation}
where $\Omega_{anc}$ and $\Omega_i$ denote the image domains of the anchor and auxiliary frames, respectively.
A larger overlap indicates that the auxiliary frame shares more scene content with the anchor frame, providing a stronger basis for cross-frame information aggregation.

However, a high overlap alone does not necessarily imply that an auxiliary frame provides useful complementary information. 
In particular, frames captured from nearly identical viewpoints may contain highly redundant observations. 
To account for cross-view variations, we further compute the average absolute difference between each aligned auxiliary and the anchor among overlap:
\begin{equation}
	d_i=\text{Avg}\left(\left|H_{i\rightarrow anc}(I_i(x))-I_{anc}(x)\right|\right),
	\label{Eq:Eq2}
\end{equation}
where $x$ denotes the pixel coordinate.
The anchor-aware score for each auxiliary frame is then defined as:
\begin{equation}
	s_{i\rightarrow anc}=o_i\cdot d_i.
\end{equation}

This formulation favors auxiliary frames that maintain sufficient geometric overlap with the anchor frame while exhibiting noticeable observation differences.
We rank all candidate frames in descending order (re-index the frames accordingly) and select the highest-scoring frame as the first auxiliary frame $S_1$.

\subsubsection{Complementarity-aware selection}
Although the first step identifies an informative auxiliary frame, directly selecting the top-ranked candidates may introduce substantial redundancy, as multiple frames may provide similar observations.
Therefore, the second step explicitly considers the complementarity between candidate frames and the already selected frame $S_1$.

For each remaining candidate, we compute its observation difference from $S_1$ after warping both frames:
\begin{equation}
	d_{i\rightarrow1}=\text{Avg}(|H_{i\rightarrow anc}(I_i(x))-H_{1\rightarrow anc}(S_1(x))|).
	\label{Eq:Eq4}
\end{equation}

We then define the complementarity-aware score as:
\begin{equation}
	s_{i\rightarrow 1}=d_{i\rightarrow 1}\cdot s_{i\rightarrow anc},
\end{equation}
By incorporating the anchor-aware score $s_{i\rightarrow anc}$, this criterion ensures the second auxiliary frame remains informative with respect to the anchor, while the inter-frame difference $d_{i\rightarrow 1}$ encourages complementary observations rather than redundant ones.
We select the candidate with the highest $s_{i\rightarrow 1}$ as $S_2$.
By iteratively repeating this step, our frame sampler can be readily extended to select $K$ frames.

It is worth noting that the proposed frame sampler is a test-time technique.
During the training phase, we instead randomly select auxiliary frames to expose the DiT model to diverse cross-view observations, thereby improving its generation capability and generalization across different frame combinations.

%% file: sec/5_experiment.tex
\begin{table*}[t]
    \centering
    \footnotesize
    \caption{Benchmark results of various competitors on the MulVi-testing and collected datasets. The best is highlighted in bold.}
    \vspace{-5pt}
    \label{tab:benchmark}
    \begin{adjustbox}{max width=\textwidth}
        \begin{tabular}{c|c|c|
                *{2}{w{c}{26pt}w{c}{35pt}w{c}{33pt}|}
                w{c}{26pt}w{c}{35pt}w{c}{33pt}}
            \toprule[1.2pt]
            \multirow{2}{*}{Method}
            & \multirow{2}{*}{Type}
            & \multirow{2}{*}{Venues}
            & \multicolumn{3}{c|}{MulVi-testing (w/ GT)}
            & \multicolumn{3}{c|}{MulVi-testing (w/o GT)}
            & \multicolumn{3}{c}{Collected Dataset} \\
            & & & PSNR$\uparrow$ & SSIM$\uparrow$ & LPIPS$\downarrow$
            & MUSIQ$\uparrow$ & CLIP-IQA$\uparrow$ & MANIQA$\uparrow$
            & MUSIQ$\uparrow$ & CLIP-IQA$\uparrow$ & MANIQA$\uparrow$ \\
            \midrule\midrule
            DSIT \cite{Hu2024NeurIPS-DSIT} & \multirow{6}{*}{SIRR} & NeurIPS'24
            & 22.20 & 0.8505 & 0.2089
            & 68.82 & 0.5467 & 0.4295
            & 60.88 & 0.4532 & 0.3316 \\
            RDNet \cite{Zhao2025CVPR-RDNet} & & CVPR'25
            & 22.34 & 0.8512 & 0.2033
            & 68.29 & 0.5272 & 0.4271
            & 60.50 & 0.4590 & 0.3281 \\
            DAI \cite{Hu2026AAAI-DAI} & & AAAI'26
            & 23.33 & 0.8565 & 0.2184
            & 69.64 & 0.5404 & 0.4409
            & 62.39 & 0.4674 & 0.3490 \\
            GFRRN \cite{Chen2026CVPR-GFRRN} & & CVPR'26
            & 22.91 & 0.8597 & 0.1921
            & 69.55 & 0.5551 & 0.4400
            & 61.78 & 0.4900 & 0.3465 \\
            GenSIRR \cite{Li2026CVPR-GenSIRR} & & CVPR'26
            & 24.01 & 0.8381 & 0.2030
            & 70.04 & 0.5572 & 0.4302
            & 64.04 & 0.5110 & 0.3578 \\
            FUMO \cite{Xu2026ECCV-FUMO} & & ECCV'26
            & 23.27 & 0.8597 & 0.1995
            & 68.61 & 0.5233 & 0.4199
            & 61.78 & 0.4686 & 0.3272 \\
            \midrule
            MIRM \cite{Li2021TIP} & \multirow{5}{*}{MFRR} & TIP'21
            & 20.14 & 0.7266 & 0.3400
            & 62.18 & 0.4794 & 0.3431
            & 56.41 & 0.4002 & 0.2904 \\
            NIR \cite{Nam2022ECCV-NIR} & & ECCV'22
            & 15.91 & 0.6409 & 0.5303
            & 42.52 & 0.3850 & 0.2695
            & 42.80 & 0.3331 & 0.2530 \\
            NSF \cite{Chugunov2024CVPR-NSF} & & CVPR'24
            & 16.12 & 0.6529 & 0.5250
            & 51.24 & 0.3476 & 0.2616
            & 46.64 & 0.3390 & 0.2180 \\
            DOVE \cite{Chen2025NIPS-DOVE} & & NeurIPS'25
            & 22.37 & 0.8361 & 0.2106
            & 69.15 & 0.5225 & 0.4229
            & 63.34 & 0.4358 & 0.3393 \\
            FlashVSR \cite{Zhang2026CVPR-FlashVSR} & & CVPR'26
            & 21.70 & 0.8412 & 0.2122
            & 69.11 & 0.5129 & 0.4191
            & 61.25 & 0.4674 & 0.3320 \\
            \midrule
            \rowcolor{MetricBlue!30}
            Ours & MFRR & \textemdash
            & \textbf{25.24}
            & \textbf{0.8761}
            & \textbf{0.1528}
            & \textbf{70.06}
            & \textbf{0.5841}
            & \textbf{0.4411}
            & \textbf{64.17}
            & \textbf{0.5112}
            & \textbf{0.3600} \\
            \bottomrule[1.2pt]
        \end{tabular}
    \end{adjustbox}
    \vspace{-10pt}
\end{table*}

\begin{figure*}[t]
	\centering
	\includegraphics[width=0.99\linewidth]{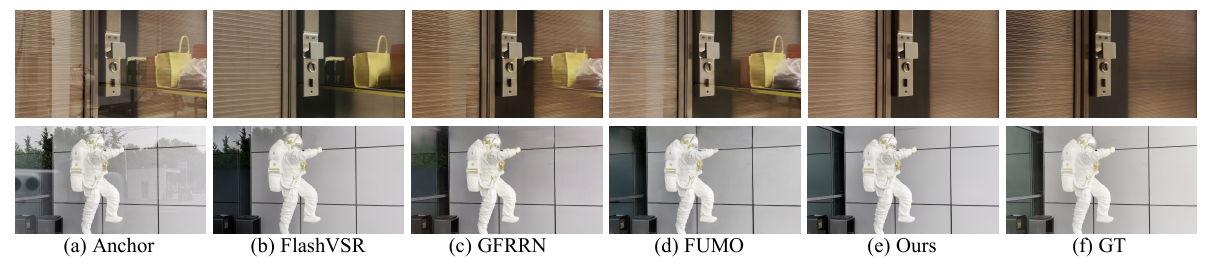}
	\vspace{-5pt}
	\caption{Visual comparisons between our MFRR-DiT and several state-of-the-art methods on samples from MulVi-testing (w/ GT). Please zoom-in on screen for more details.}
	\vspace{-10pt}
	\label{fig:fig6}
\end{figure*}

\begin{figure*}[!h]
	\centering
	\includegraphics[width=0.99\linewidth]{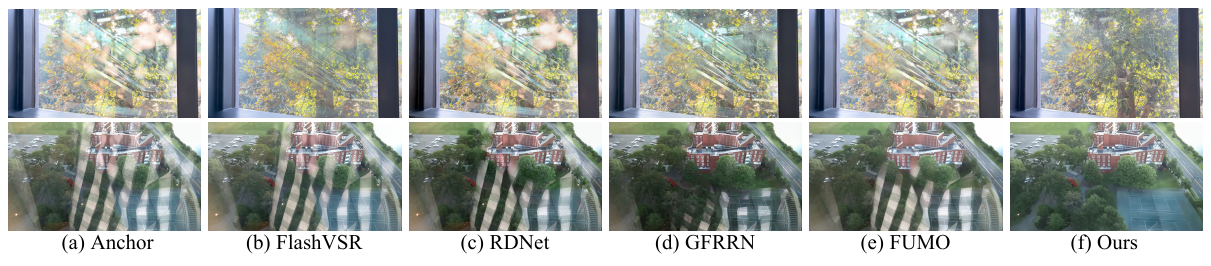}
	\vspace{-5pt}
	\caption{Visual comparisons between our MFRR-DiT and several state-of-the-art methods on samples from MulVi-testing (w/o GT) (top row) and our collected dataset (bottom row). Please zoom-in on screen for more details.}
	\vspace{-10pt}
	\label{fig:fig7}
\end{figure*}

\section{Experiments}
\label{sec: experiments}
\subsection{Implementation details}
\label{subsec: implementation}
\noindent\textbf{Dataset.}
Our MFRR-DiT is mainly trained and evaluated on our MulVi dataset.
For training the pre-adaptation stage, we randomly sample 5,000 synthetic pairs from the PASCAL VOC dataset \cite{Everingham2010IJCV-PascalVOC} in each epoch.
90 image pairs from ``Real'' dataset \cite{Zhang2018CVPR} and 200 image pairs from ``Nature'' dataset \cite{Li2020CVPR-IBCLN} are also included for training.
We also employ the RRW \cite{Zhu2024CVPR-RRW} to augment our training data by applying simple geometric transformations to warp image sequences to other viewpoints. 
\textbf{More details can be found in our supplementary material.}
For evaluating the performance, we further collect some testing sequences from previous works \cite{Chugunov2024CVPR-NSF,Xue2015TOG,Li2013ICCV,Guo2022CVPR-nerfren,Zhu2022NeurIPS}.

\noindent\textbf{Training details.}
We implement our MFRR-DiT in PyTorch with 8 NVIDIA A800 GPUs.
In both training stages, we fine-tune the DiT backbone while keeping the VAE frozen.
Stage I is initialized from the pre-trained FLUX.2 Klein 4B model \cite{flux-2-2025}, and its learned weights are subsequently used to initialize Stage II.
Both stages are trained for 2,000 iterations using the AdamW optimizer \cite{Loshchilov2019ICLR-AdamW,Kingma2015ICLR-Adam}, with an effective batch size of 80 and a patch size of $512\times 512$.
The cosine annealing strategy \cite{Loshchilov2017-ICLR-SGDR,He2019CVPR-Bags} is adopted to decrease the learning rate from $5 \times 10^{-5}$ to $1 \times 10^{-6}$.
When training the Stage II, we set $K=2$ in our implementation.

\subsection{Performance evaluation}
\noindent\textbf{Quantitative comparison.}
We compare our MFRR-DiT with 6 SIRR methods including DSIT \cite{Hu2024NeurIPS-DSIT}, RDNet \cite{Zhao2025CVPR-RDNet}, DAI \cite{Hu2026AAAI-DAI}, GFRRN \cite{Chen2026CVPR-GFRRN}, GenSIRR \cite{Li2026CVPR-GenSIRR}, FUMO \cite{Xu2026ECCV-FUMO}, and 5 MFRR methods including MIRM \cite{Li2021TIP}, NIR \cite{Nam2022ECCV-NIR}, NSF \cite{Chugunov2024CVPR-NSF}, DOVE \cite{Chen2025NIPS-DOVE}, FlashVSR \cite{Zhang2026CVPR-FlashVSR} on our MulVi-testing and our collected dataset.
\hzw{We fine-tune the competitors on our training data and choose the better performance to report.
Note that, NIR and NSF are originally designed for burst sequences, and DOVE and FlashVSR are originally designed for super-resolution \cite{He2023PR,He2020TMM,He2019TCSVT}.}
Table~\ref{tab:benchmark} shows the benchmark results on our testing datasets, and our MFRR-DiT ranks the first in all comparisons.

\noindent\textbf{Qualitative comparison.}
Fig.~\ref{fig:fig6} shows the visual results on MulVi-testing (w/ GT).
Fig.~\ref{fig:fig7} shows the visual results on MulVi-testing (w/o GT) and our collected dataset.
\hzw{Our MFRR-DiT achieves cleaner reflection removal and better preserves background details than competing methods, particularly in regions with strong reflections.}
\textbf{Due to the limited space, additional examples can be found in our supplementary material.}

\subsection{Ablation study}
\label{subsec: ablation}

In this section, we first investigate the effectiveness of our 3D scene-based rendering pipeline for constructing MulVi-training.
Later, to validate the effectiveness of our MFRR-DiT, we conduct ablation study on the key components (\ie, multi-frame conditioning, ACA, frame sampler).

\begin{table}[t]
	\centering
	\caption{Ablation study on the training data.}
	\vspace{-5pt}
	\label{tab:data_ablation}
	\footnotesize
	\begin{tabular}{c|ccc|cc}
		\toprule[1.2pt]
		Setting&2D-based  & MulVi-training & RRW & PSNR& SSIM \\
		\midrule\midrule
		A& $\checkmark$ & $\times$ & $\times$& 23.64 & 0.8601 \\
		B & $\times$ & $\checkmark$ & $\times$& 24.55 & 0.8678 \\
		\rowcolor{MetricBlue!30}C (Ours)& $\times$ & $\checkmark$ & $\checkmark$& 25.24 & 0.8761 \\
		\bottomrule[1.2pt]
	\end{tabular}
	\vspace{-10pt}
\end{table}

\noindent\textbf{Effectiveness of our MulVi-training.}
Our MulVi-training is rendered based on synthetic 3D scenes, providing more accurate geometric consistency across different viewpoints.
This advantage is further supported by our experimental results, which show a 0.91 dB PSNR improvement over the 2D image-based synthesis baseline (see in Table~\ref{tab:data_ablation}).
In addition, incorporating some real-world sequences from RRW \cite{Zhu2024CVPR-RRW} further improves the performance.

\noindent\textbf{Multi-frame conditioning.}
We first examine the effectiveness of multi-frame conditioning over single-frame input.
Specifically, we employ the single-frame setting as the baseline (\ie, Stage I-only).
Subsequently, we progressively increase the number of auxiliary frames $K$ to investigate how additional viewpoints affect reflection removal performance.

As shown in Table~\ref{tab:tab3}, incorporating auxiliary frames consistently improves reflection removal performance over the single-frame baseline.
Specifically, using a single auxiliary frame improves PSNR from 23.04 dB to 24.92 dB, while increasing the number of auxiliary frames to two further boosts PSNR to 25.24 dB, achieving an overall gain of 2.20 dB.
It demonstrates the effectiveness of leveraging complementary observations from different viewpoints.
Interestingly, further increasing the number of auxiliary frames does not yield obvious gains.
This suggests that the effectiveness of multi-frame conditioning depends not only on the number of auxiliary frames but also on their relevance and complementarity.

\begin{table}[h]
	\footnotesize
	\caption{Ablation study on multi-frame conditioning and the number of $K$.}
	\vspace{-5pt}
	\label{tab:tab3}
	\centering
	\adjustbox{width=\linewidth}{
		\begin{tabular}{c|ccccc}
			\toprule[1.2pt]
			Setting &Stage I-only&$K=1$&\cellcolor{MetricBlue!30}$K=2$&$K=3$&$K=4$\\
			\midrule\midrule
			PSNR & 23.04  &24.92&\cellcolor{MetricBlue!30}25.24&25.15&25.18\\
			SSIM & 0.8490  &0.8720&\cellcolor{MetricBlue!30}0.8761&0.8751&0.8750\\
			\bottomrule[1.2pt]
		\end{tabular}
	}\vspace{-10pt}
\end{table}

\noindent\textbf{Anchor-centric attention.}
The proposed Anchor-Centric Attention (ACA) consists of two key components: asymmetric attention and RoPE offset. 
As shown in Table~\ref{tab:ACA}, removing either component results in degraded performance, demonstrating the importance of both designs. 
Meanwhile, incorporating either component individually improves performance over the baseline with the original FLUX.2 self-attention mechanism. 
These results highlight the individual effectiveness and complementary benefits of the two components.

\begin{table}[h]
	\centering
	\caption{Ablation study on key components of proposed ACA.}
	\vspace{-5pt}
	\label{tab:ACA}
	\footnotesize
	\begin{tabular}{c|cc|cc}
		\toprule[1.2pt]
		Setting
		& RoPE offset& Asymmetric attention& PSNR& SSIM \\
		\midrule\midrule
		A& $\times$ & $\times$& 24.97 & 0.8740 \\
		B& $\checkmark$ & $\times$& 25.11 & 0.8750 \\
		C& $\times$ & $\checkmark$& 25.14 & 0.8745 \\
		\rowcolor{MetricBlue!30}D (Ours)& $\checkmark$ & $\checkmark$& 25.24 & 0.8761 \\
		\bottomrule[1.2pt]
	\end{tabular}
	\vspace{-10pt}
\end{table}

\begin{figure}[h]
	\centering
	\includegraphics[width=0.9\linewidth]{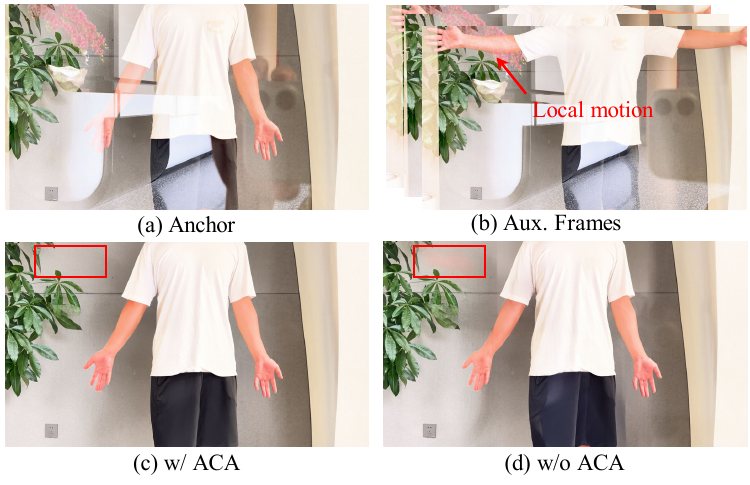}
	\vspace{-5pt}
	\caption{Visual results of our MFRR-DiT with and without ACA under local motion.}
	\vspace{-10pt}
	\label{fig:fig8}
\end{figure}

As shown in Fig.~\ref{fig:fig8}, ACA helps mitigate interference caused by local motion in auxiliary frames.
Without ACA, the restored image exhibits noticeable arm-shaped artifacts in the highlighted region.
In contrast, incorporating ACA effectively suppresses these artifacts and produces a more coherent reconstruction.
We attribute this improvement to the anchor-centric interaction scheme, which restricts direct information exchange among auxiliary frames and thereby mitigates the propagation of inconsistent information caused by local motion.

\noindent\textbf{Frame sampler.}
Although the frame sampler is a simple test-time technique, it plays a crucial role in our framework. 
First, it enables MFRR-DiT to accommodate a variable number of candidate frames, overcoming the constraint on the number of input frames. 
Second, it automatically selects the informative and mutually complementary auxiliary frames for effective cross-view information aggregation.

\begin{table}[h]
	\footnotesize
	\caption{Ablation study on frame sampler.}
	\vspace{-5pt}
	\label{tab:frame sampler}
	\centering
	\adjustbox{width=\linewidth}{
	\begin{tabular}{c|cccc}
		\toprule[1.2pt]
		Setting & Random (worst) & Random (best)&+ anchor-aware& \cellcolor{MetricBlue!30}Ours \\
		\midrule\midrule
		PSNR & 24.86 & 25.00 & 25.12 & \cellcolor{MetricBlue!30}25.24 \\
		SSIM & 0.8636 & 0.8737 & 0.8749 & \cellcolor{MetricBlue!30}0.8761\\
		\bottomrule[1.2pt]
	\end{tabular}
	}\vspace{-10pt}
\end{table}

Random sampling arbitrarily selects two auxiliary frames without considering their informativeness and complementarity, leading to considerable performance variations across different selections. 
We conduct 10 independent random sampling trials and report the best and worst results in Table~\ref{tab:frame sampler}.
By employing the step 1 (\ie, anchor-aware selection), an improvement is observed.
Notably, our two-step selection strategy outperforms even the best-performing random selection, demonstrating its effectiveness in identifying suitable auxiliary frames. 
The reported best result is obtained from only 10 random trials and does not necessarily represent the globally optimal selection.

\subsection{User study}

\begin{figure}[h]
	\centering
	\includegraphics[width=0.85\linewidth]{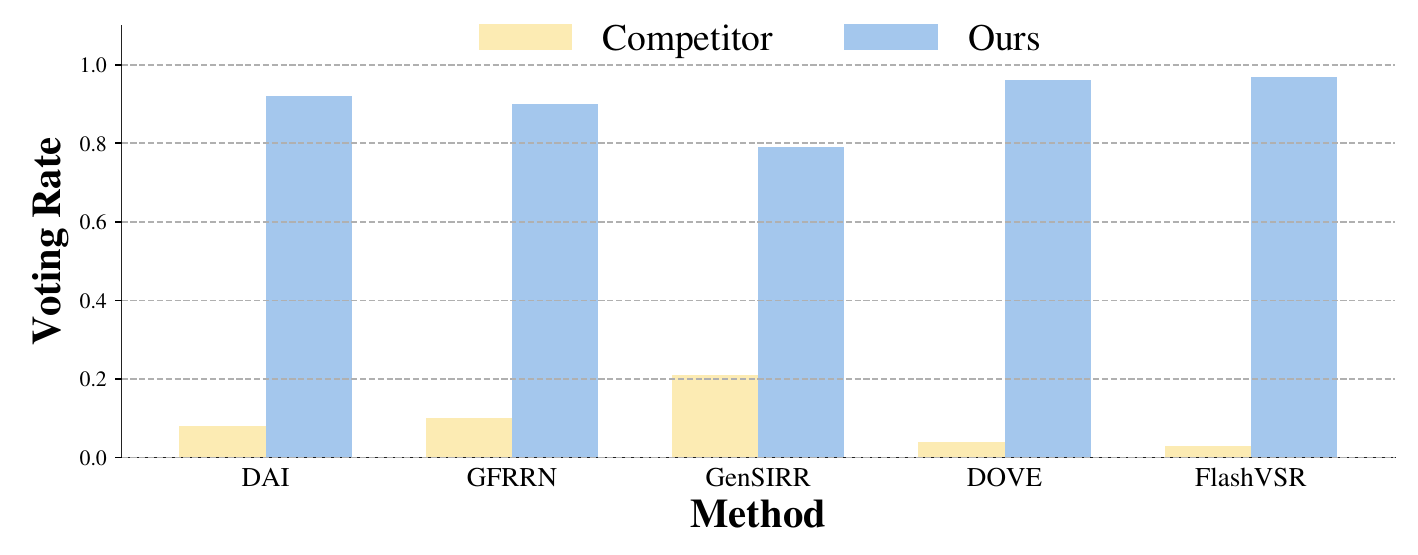}
	\vspace{-5pt}
	\caption{User study on our MulVi-testing (w/o GT).}
	\vspace{-10pt}
	\label{fig:fig9}
\end{figure}

We conduct a user study with ten image processing experts on all images in MulVi-testing (w/o GT). For every image, the experts compare the MFRR-DiT result with five alternatives in turn, choosing the visually better result after at least 10 seconds per pair. We report the preference rates in Fig. \ref{fig:fig9}, which demonstrate that MFRR-DiT is generally favored over comparative methods.

%% file: sec/6_conclusion.tex
\section{Conclusion}
\label{sec: conclusion}

We revisit MFRR and introduce MulVi, a dataset comprising 3D-rendered training sequences and real-world evaluation benchmarks. 
We further propose MFRR-DiT, an end-to-end DiT-based framework that effectively exploits complementary information across multiple viewpoints without explicit geometric alignment. 
To facilitate cross-frame information aggregation, we develop Anchor-Centric Attention (ACA) to regulate information exchange and mitigate interference from local motion, along with a test-time frame sampler to select informative and complementary auxiliary frames. 
Extensive experiments demonstrate the effectiveness of our approach.